\documentclass[sigconf]{acmart}
\renewcommand\footnotetextcopyrightpermission[1]{}
\usepackage{pdfpages}
\usepackage{graphicx} 
\usepackage{booktabs}
\usepackage{multirow}
\usepackage{setspace}
\usepackage{color}
\AtBeginDocument{%
  }

\setcopyright{acmlicensed}
\copyrightyear{2023}
\acmYear{2023}
\acmDOI{XXXXXXX.XXXXXXX}

\acmConference[Conference acronym 'ICMSSP]{Make sure to enter the correct
  conference title from your rights confirmation emai}{May 24--26,
  2024}{Bangkok, Thailand}
\acmISBN{979-8-4007-1691-1/24/05}

\begin{document}

\title{Improve Cross-Architecture Generalization on Dataset Distillation}

\author{Binglin Zhou}
\authornote{All authors contributed equally to this research.}
\email{zhoubinglin@sjtu.edu.cn}
\affiliation{%
  \institution{Shanghai Jiao Tong University}
  \city{Shanghai}
  \country{China}
}
\author{Linhao Zhong}
\authornotemark[1]
\email{zhongzero@sjtu.edu.cn}
\affiliation{%
  \institution{Shanghai Jiao Tong University}
  \city{Shanghai}
  \country{China}
}
\author{Wentao Chen}
\authornotemark[1]
\email{leonard_chen@sjtu.edu.cn}
\affiliation{%
  \institution{Shanghai Jiao Tong University}
  \city{Shanghai}
  \country{China}
}


\begin{abstract}
{
\setstretch{1.25}
Dataset distillation, a pragmatic approach in machine learning, aims to create a smaller synthetic dataset from a larger existing dataset. However, existing distillation methods primarily adopt a model-based paradigm, where the synthetic dataset inherits model-specific biases, limiting its generalizability to alternative models. In response to this constraint, we propose a novel methodology termed "model pool". This approach involves selecting models from a diverse model pool based on a specific probability distribution during the data distillation process. Additionally, we integrate our model pool with the established knowledge distillation approach and apply knowledge distillation to the test process of the distilled dataset. Our experimental results validate the effectiveness of the model pool approach across a range of existing models while testing, demonstrating superior performance compared to existing methodologies. Our code can is available at: \href{https://github.com/Distill-Generalization-Group/Distill-Generalization}{https://github.com/Distill-Generalization-Group/Distill-Generalization}

}
\end{abstract}



\maketitle

\section{Introduction}
{\Large 
\setstretch{1.25}
As the structural complexity of neural network models continues to advance, deep learning has achieved remarkable success across diverse domains. However, the pursuit of enhanced performance in deeper networks needs a commensurate increase in the quantity of training data, resulting in a significant slowdown in training speed. Addressing this challenge, one viable approach involves reducing the size of the training dataset. Proposed by Wang et al., Dataset Distillation\cite{wang2018dataset}(also called \textbf{DD}) emerges as an effective method to achieve dataset size reduction while ensuring the efficacy of network training. The core objective of dataset distillation is the synthesis of a substantially smaller dataset derived from the original dataset, and a model trained on this synthesized dataset still exhibits robust performance on the original test dataset.

\textbf{DD} represents a category of meta-learning-based dataset distillation methods. Since the introduction of \textbf{DD}, an increasing array of dataset distillation methods has been proposed, categorized into various types such as gradient matching-based (Zhao et al. \cite{zhao2020dataset}), trajectory matching-based (Cazenavette et al. \cite{cazenavette2022dataset}), and distribution matching-based (Zhao et al. \cite{zhao2022dataset}), among others. The classification of dataset distillation methods proposed by Sachdeva et al. \cite{sachdeva2023data} encompasses four distinct categories, as depicted in Table \ref{dataset_distillation_methods}.

}
\begin{table}[h]  
    \centering  

    \begin{tabular}{|c|c|}  
        \hline  
        \textbf{Type} & \textbf{Name} \\  
        \hline
        \textbf{Meta Model Learning} & DD, KIP, FRePO etc. \\  
        \hline  
        \textbf{Gradient Matching} & DC, DSA, DCC etc.  \\  
        \hline
        \textbf{Trajectory Matching} & MTT, HaBa, TESLA etc.\\  
        \hline
        \textbf{Distribution Matching} & DM, CAFE, KFS etc.  \\  
        \hline  
    \end{tabular}
    \caption{dataset distillation methods}  
    \label{dataset_distillation_methods}

\end{table}

{\Large
\setstretch{1.25}
The prevailing dataset distillation methods predominantly adhere to a model-based paradigm, wherein a specific model architecture is chosen to derive the final distilled dataset. Notably, most meta-learning, gradient matching, and trajectory matching-based approaches are model-centric, aiming to iteratively update both the distilled dataset and model parameters within their respective architectures. In contrast, \textbf{DM}\cite{zhao2022dataset}, a classical distribution matching-based dataset distillation method is model-agnostic. This implies that \textbf{DM} does not seek to enhance distillation within a specific model's architecture but tries to directly align the distribution of the distilled dataset with that of the original dataset.

Intuitively, datasets generated through model-based methods show superior performance when tested on the architecture employed during the distillation process. However, a noticeable reduction in performance occurs when these datasets are assessed on alternative model architectures—a phenomenon we call as "performance reduction in cross-architecture." \cite{cui2022dcbench} This raises the critical question of how to endow dataset distillation methods with the capability of cross-architecture generalization. 

Afterward, some research was conducted to explore the generalization ability of distilled datasets. In particular, a paper \cite{zhong2023mitigating} proposed a method called DropPath to improve the generalization ability of \textbf{MTT}\cite{cazenavette2022dataset} and \textbf{FRePO}\cite{zhou2022dataset}. 

In this study, we present a novel dataset distillation method designed to enhance cross-architecture generalization, building upon \textbf{DC} \cite{zhao2020dataset}, a gradient matching-based approach. First, our method introduces the concept of "model pool", which is used to promote cross-architecture generalization ability and can be easily extensible to other existing model-based dataset distillation methods. Also, drawing inspiration from knowledge distillation, we leverage a similar strategy when deploying our distilled dataset on alternative model architectures. Experimental evaluations conducted on the CIFAR-10 dataset demonstrate that our proposed method outperforms the original DC method in cross-architecture generalization.

}

\section{Related Work}
{\Large
\setstretch{1.25}
Our work is based on dataset distillation, model generalization, and knowledge distill. In this section, we will introduce the related work of them respectively.

}
\subsection{Dataset Distillation}
{\Large
\setstretch{1.25}
Dataset distillation, a process aimed at deriving a downscaled dataset from an original dataset while preserving comparable results during model training, has garnered considerable attention, leading to the development of numerous methodologies. Notably, meta-learning-based approaches constitute a category of dataset distillation methods. For instance, the pioneering \textbf{DD}\cite{wang2018dataset} method updates model parameters in the inner loop and refines the distilled dataset in the outer loop, mirroring the paradigm of the MAML\cite{finn2017model} method.

Another class of dataset distillation methods centers around gradient matching, with the seminal contribution of \textbf{DC} paving the way for subsequent advancements. Building upon this foundation, \textbf{IDC}\cite{kim2022dataset} extends the gradient matching framework. In a parallel vein, \textbf{MTT} introduces a dataset distillation method grounded in trajectory matching.

Diverging from these approaches, distribution matching represents a distinct dataset distillation methodology, as exemplified by \textbf{DM}. This method deviates from the model-centric focus by directly aligning the distribution of the distilled dataset with that of the original dataset. The diverse range of dataset distillation techniques underscores the evolving landscape of methodologies, each with its unique strengths and applications.

}
\subsection{Model Generalization}
{\Large
\setstretch{1.25}
Addressing model generalization is a common concern in machine learning, where the expectation is that a model performs effectively on the test set following training on the training set. Numerous methodologies have been proposed to enhance model generalization.

One category of methods centers on model ensemble techniques. Model ensemble involves integrating multiple models to enhance overall generalization. In classification tasks, popular methods include bagging and boosting. Bagging entails training multiple models on distinct subsets of the training set, amalgamating them to produce the final model, as exemplified by random forest\cite{Breiman2001RandomF}. Conversely, boosting involves training multiple models on the same training set, adjusting the weights of examples based on the performance of the preceding training epochs, as illustrated by AdaBoost\cite{10.1007/3-540-59119-2_166}.

Building upon the principles of model ensemble, we introduce the \textbf{model pool} method to augment the generalization capabilities of the distilled dataset.

}
\subsection{Knowledge Distillation}
{\Large
\setstretch{1.25}
Knowledge distillation is a method to transfer knowledge from a large model to a small model. It's first proposed in this paper \cite{hinton2015distilling}. The large model is called the teacher model and the small model is called the student model. The student model is trained to mimic the output of the teacher model. The student model is usually a shallow neural network with fewer parameters and faster inference speed. The teacher model is usually a deep neural network with more parameters and slower inference speed. The student model is trained on the same dataset as the teacher model to minimize the difference between the output of the student model and the output of the teacher model (i.e. maintain the knowledge as much as possible). 

After knowledge distillation was proposed, a lot of work has been put into exploring further applications in diverse fields, such as speech recognition, image recognition, and natural language processing. It proved that knowledge distillation is a great success in these various fields \cite{Gou_2021}.

However, there are some arguments against knowledge distillation. A paper \cite{stanton2021does} said that, while knowledge distillation improves the generalization of student models, there often remains a surprisingly large discrepancy between the predictive distributions of the teacher and the student, even in cases when the student can perfectly match the teacher. 

}
\section{Methods}

{\Large
\setstretch{1.25}
In this section, we will introduce the methods we used to perform dataset distillation and to mitigate the performance reduction in cross-architecture tests.

\subsection{Preliminary: Dataset Distillation Based On Gradient Matching}

Before the generalization part, we use gradient matching \cite{zhao2020dataset} to complete the data distillation process. The basic idea of gradient matching is to minimize the distance between the gradients of the loss function of the distilled dataset and the original dataset. The objective function of the gradient matching method is as follows:

}
\begin{align*}
&\underset{\mathcal{D}_{\text {syn }}}{\arg \min } \underset{\substack{\theta_0 \sim \mathbf{P}_\theta \\ c \sim \mathcal{C}}}{\mathbb{E}}\left[\sum_{t=0}^T \mathbf{D}\left(\nabla_\theta \mathcal{L}_{\mathcal{D}^c}\left(\theta_t\right), \nabla_\theta \mathcal{L}_{\mathcal{D}_{\text {syn }}^c}\left(\theta_t\right)\right)\right] \quad \\
&\text { s.t. } \quad \theta_{t+1} \leftarrow \theta_t-\eta \cdot \nabla_\theta \mathcal{L}_{\mathcal{D}_{\text {syn }}}\left(\theta_t\right)
\end{align*}
{\Large
\setstretch{1.25}
Where $\mathcal{D}_{\text{syn}}$ denotes the distilled dataset, $\mathcal{D}^c$ and $\mathcal{D}_{\text{syn}}^c$ denote the $c$-th class of the original dataset and the distilled dataset respectively, $\theta$ denotes the parameters of the model(i.e. the model we choose to finish dataset distillation), $\mathcal{L}_{\mathcal{D}}$ denotes the loss function on the training set $\mathcal{D}$, $\eta$ denotes the learning rate, $T$ denotes the the number of steps we will match in the training process. And, $\mathbf{D}: \mathbb{R}^{|\theta|} \times \mathbb{R}^{|\theta|} \mapsto \mathbb{R}$ is a distance function, which is usually chosen as the cosine distance. 

Then we will introduce the method we used to improve the generalization ability of the distilled dataset in the next two subsections.

}
\subsection{Model Pool}\label{model_pool}
\begin{figure*}[h]
\centering
\includegraphics[width=\linewidth]{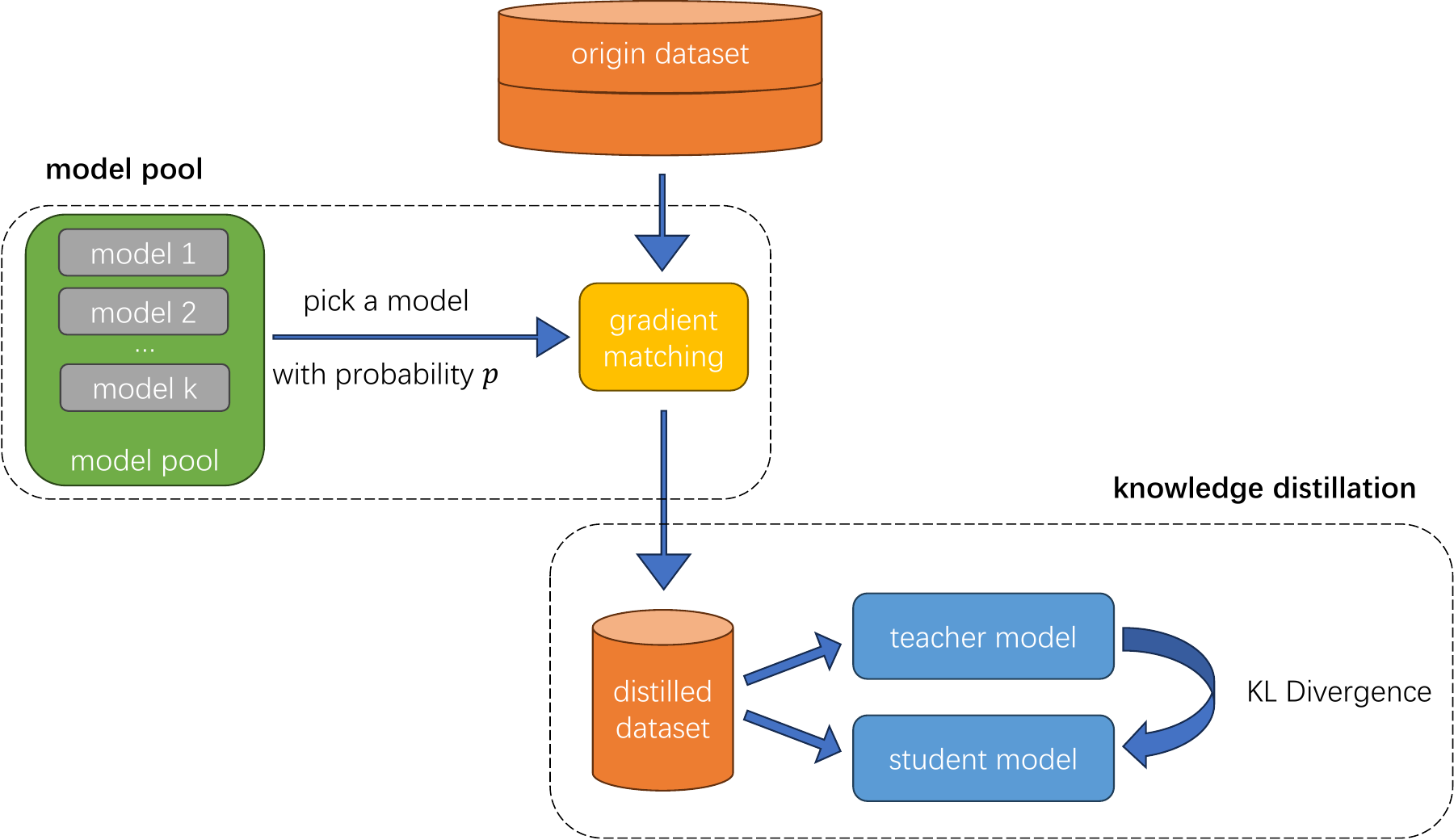}
\caption{Overview of model pool \ref{model_pool} and knowledge distillation \ref{knowledge distillation} method. The model pool is a set of different models with different architectures. Every model in the model pool has its own probability to be chosen. Knowledge distillation is performed on the distilled dataset. The two methods are independent. }
\label{structure}
\end{figure*}

{\Large
\setstretch{1.25}
The original gradient matching method is one kind of model-based method. It means that it will choose a specific model architecture and use only this model to obtain the final distilled dataset. This kind of method will obtain a great performance when testing on the prior specific model that it used in the dataset distillation process, but it will get a reduction in performance when testing on other model architectures. This is not the result we expected. In contrast, we want our distilled dataset to have a better performance in cross-architecture generalization.

We propose the concept of model pool. The model pool is a set of different models with different architectures. Every model in the model pool has its own probability to be chosen. In the original model-based method, we may reinitialize the parameters of this model many times in the total training process, noting that the model architecture doesn't change yet. In our method, the model we use and reinitialize each time may be different and it is chosen from the model pool with a given probability. In this way, we can make our distilled dataset more general as it is not only trained on a specific architecture of the model but trained on multiple architectures of models in the model pool. Our model pool can be visualized in Fig. \ref{structure}. 

}

{\Large
\setstretch{1.25}
While the model pool can improve the generalization ability of the distilled dataset, it also brings some problems. The main problem is that it is hard to make the training process converge. In intuition, consider there is a function $f(x)$ whose inputs are the parameters of the distilled dataset and the output is the loss which can show the performance of the distilled dataset. In the original model-based method, the distilled dataset is trained on a specific model architecture and the parameters will be reinitialized many times. Each time the parameters are reinitialized, the backward propagation process will try to make the parameters of the distilled dataset to be the current local optimal parameters. However as the model architecture is specific, the total process of training can be seen as a process of finding the global optimal parameters of the distilled dataset based on the specific model architecture. It is still easy to converge. However, in our model pool method, the distilled dataset is trained on multiple architectures of models. The total training process can be seen as finding the global optimal parameters of the distilled dataset based on multiple architectures of models. These architectures of models' global optimal parameters may become extremely different, especially when the architectures of models are extremely varied. In this way, it is hard to converge. Therefore, we propose the concept of the main model. The main model is a model in the model pool and it has a much greater probability to be chosen than other models. Besides the main model, we also limit the other models in the model pool to be similar to the main model. In this way, our training process can be seen as a process of finding the global optimal parameters of the distilled dataset based on multiple architectures of models with similar global optimal parameters and they have main global optimal parameters. It is much easier to converge while the distilled dataset can still improve the generalization ability in cross-architecture.

}

\subsection{Knowledge Distillation}\label{knowledge distillation}
{\Large
\setstretch{1.25}
Knowledge distillation is a useful method used in transfer learning. Considering that the space and resources are limited, sometimes people need a smaller model (student model) that its performance (i.e. 'knowledge') is as well as the original bigger model (teacher model). The core idea behind knowledge distillation involves transferring the insights of the teacher model to the student model by aligning the student's predictions (or internal activation functions) with those of the teacher as closely as possible. Specifically, the KL divergence, utilized with a certain temperature, acts as a mechanism to align the predictions of the student and teacher models. This KL divergence is then incorporated as a regularization term in conjunction with the classification loss (Cross-Entropy Loss). Mathematically, the total loss function is expressed as follows:

}
\begin{align}
    \mathcal{L}\left(\mathbf{y}_s, \mathbf{y}_t, y\right)=\mathcal{L}_{K L}\left(\mathbf{y}_s, \mathbf{y}_t\right) \cdot \alpha \cdot \tau^2+\mathcal{L}_{C E}\left(\mathbf{y}_s, y\right) \cdot(1-\alpha)
\end{align}
{\Large
\setstretch{1.25}
where $\tau$ denotes the temperature factor, and $\alpha\in(0,1)$ denotes the weight factor to balance the KL divergence $\mathcal{L}_{KL}$ and cross-entropy $\mathcal{L}_{CE}$ . The output logits of the student model and teacher model are denoted by $\mathbf{y}_s$ and $\mathbf{y}_t$ , respectively. $y$ denotes the target.

In this background, we will perform knowledge distillation on the distilled dataset. In other words, both the teacher model and student model are trained on the distilled data, and then we try to minimize the difference (KL divergence) between the two models and their classification loss (cross-entropy loss). The teacher model is fixed as ConvNet while the student can be various. We can test the generalization ability of different student models by directing their knowledge to the fixed teacher model as closely as possible. 

}

\section{Experiments}

{\Large
\setstretch{1.25}
In this section, we will introduce the experiment we conducted to verify the effectiveness of our method. In Section \ref{section:settings}, we will introduce the settings of our experiment. In Section \ref{section:Result Analysis}, we will introduce the result of our experiment and analyze the result.

}

\subsection{Settings}\label{section:settings}
{\Large
\setstretch{1.25}
We conduct the experiment on CIFAR10 dataset. The CIFAR10 dataset consists of $60000$ $32\times 32$ colour images in $10$ classes, with $6000$ images per class. There are $50000$ training images and $10000$ test images. We use the training set as the original dataset to do dataset distillation and test our synthesized dataset on the origin test dataset. We use the gradient matching method based on $3$-layer ConvNet to distill the data as our baseline method. 

For each method(baseline, only model pool, only knowledge distill, both model pool and knowledge distill), we set IPC(Images Per Class) as $1, 10$. Then for each IPC, we set the iteration number in gradient matching as $1000, 2000 \text{and} 3000$. The total result is shown in Table \ref{result}.

}

\subsubsection{Model Pool}

{\Large
\setstretch{1.25}
The model pool method can promote the generalization ability of the distilled dataset, but extremely various models in the model pool will make the training process hard to converge as we have mentioned before. Therefore, in our experiment, we choose ConvNet as the main model with a probability of 0.9 and this main ConvNet model's specific architecture has been shown in Table \ref{MainConvNet}. Besides the main model, we also choose other models in the model pool with a total probability of 0.1. To make the training process converge, the other models in the model pool are also ConvNet models with some differences in the specific architecture. We randomly choose the type of network activation function, the type of normalization layer and the type of pooling layer in a certain range. Following the original gradient matching method \textbf{DC}, we set the model's network width as 128 and network depth as 3. The specific architecture of the other models in the model pool has been shown in Table \ref{RandomConvNet}. We also do some experiments on the model pool method setting on Table \ref{model_pool_ablation}, which will demonstrate the effectiveness of the idea of model pool and the similarity of the other models in the model pool.

}

\begin{table}[h]

    \begin{tabular}{p{3cm}|p{4cm}}
      \toprule
      hyper parameter & specific setting \\
      \midrule
      net width & 128 \\
      net depth & 3 \\
      net activation layer & relu \\
      net normalization layer & instancenorm \\
      net pooling layer & avgpooling \\
      \bottomrule
    \end{tabular}
    
    \caption{Main ConvNet in model pool. The main ConvNet model is chosen with a probability of 0.9. The specific architecture of the main ConvNet model is shown in this table, including the net width, net depth, net activation layer, net normalization layer and net pooling layer.}
    \label{MainConvNet}
\end{table}

\begin{table}[h]

    \begin{tabular}{p{3cm}|p{4cm}}
      \toprule
      hyper parameter & specific setting \\
      \midrule
      net width & 128 \\
      net activation layer & \{relu, sigmoid, leakyrelu, swish\} \\
      net normalization layer & \{instancenorm, batchnorm, layernorm, groupnorm\} \\
      net pooling layer & \{avgpooling, maxpooling\} \\
      \bottomrule
    \end{tabular}
    
    \caption{Other Random ConvNet in model pool. The other ConvNet models are chosen with a total probability of 0.1. The specific architecture of the other ConvNet models is shown in this table, including the net width, net depth, net activation layer, net normalization layer, and net pooling layer. The net width and net depth remain the same in all these other ConvNet models. The net activation layer, net normalization layer, and net pooling layer are randomly chosen from the given range with the same probability, which is shown in this table.}
    \label{RandomConvNet}
\end{table}

\begin{table*}[ht]
  \centering  

  \begin{tabular}{|c|c|c|c|c|c|c|c|c|}  
  \hline
  \textbf{ipc} & \textbf{iteration} & \textbf{method}  & \textbf{ConvNet} & \textbf{MLP} & \textbf{LeNet} & \textbf{AlexNet} & \textbf{VGG11} & \textbf{ResNet18}  \\
  \hline
  \multirow{9}{*}{1} & \multirow{3}{*}{1000} & baseline & 28.49 & 26.59 & 20.97 & 21.83 & 26.53 & 14.30 \\
  & & w/ model pool & - & 27.88 & 20.75 & 21.61 & 26.51 & 15.17 \\
  & & w/ KD & - & 26.52 & 20.87 & 22.07 & 25.25 & 18.69 \\
  & & w/ model pool \& KD & - & \textbf{\textcolor{red}{28.17}} & \textbf{\textcolor{red}{21.22}} & \textbf{\textcolor{red}{22.54}} & \textbf{\textcolor{red}{27.15}} & \textbf{\textcolor{red}{22.03}} \\
  \cline{2-9}
  & \multirow{3}{*}{2000} & baseline & 28.62 & 26.59 & 20.91 & 21.36 & 26.23 & 14.68 \\
  & & w/ model pool & - & 27.50 & 21.43 & 21.71 & 26.55 & 14.04 \\
  & & w/ KD & - & 27.08 & 21.03 & 21.68 & 25.36 & 19.10 \\
  & & w/ model pool \& KD & - & \textbf{\textcolor{red}{28.10}} & \textbf{\textcolor{red}{22.11}} & \textbf{\textcolor{red}{22.18}} & \textbf{\textcolor{red}{26.95}} & \textbf{\textcolor{red}{21.84}} \\
  \cline{2-9}
  & \multirow{3}{*}{3000} & baseline & 28.34 & 26.07 & 21.01 & 21.42 & 26.51 & 14.27 \\
  & & w/ model pool & - & 28.01 & 21.26 & 21.62 & 26.82 & 14.09 \\
  & & w/ KD & - & 26.88 & 21.66 & 21.74 & 25.42 & 18.97 \\
  & & w/ model pool \& KD & - & \textbf{\textcolor{red}{28.07}} & \textbf{\textcolor{red}{22.50}} & \textbf{\textcolor{red}{21.86}} & \textbf{\textcolor{red}{27.25}} & \textbf{\textcolor{red}{22.27}} \\
  \hline
  \hline
  \multirow{9}{*}{10} & \multirow{3}{*}{1000} & baseline & 44.60 & 27.35 & 22.16 & 20.64 & 30.38 & 15.60 \\
  & & w/ model pool & - & 29.29 & 23.81 & 23.40 & 30.40 & 14.94 \\
  & & w/ KD & - & 27.93 & 23.63 & 21.55 & \textbf{\textcolor{red}{35.65}} & 19.43 \\
  & & w/ model pool \& KD & - & \textbf{\textcolor{red}{30.02}} & \textbf{\textcolor{red}{24.91}} & \textbf{\textcolor{red}{25.38}} & 33.85 & \textbf{\textcolor{red}{21.51}} \\
  \cline{2-9}
  & \multirow{3}{*}{2000} & baseline & 44.53 & 27.57 & 23.71 & 20.90 & 30.03 & 15.60 \\
  & & w/ model pool & - & 29.20 & 24.67 & 22.45 & 30.20 & 15.68 \\
  & & w/ KD & - & 27.78 & 23.43 & 21.98 & \textbf{\textcolor{red}{35.53}} & 19.37 \\
  & & w/ model pool \& KD & - & \textbf{\textcolor{red}{30.23}} & \textbf{\textcolor{red}{24.96}} & \textbf{\textcolor{red}{24.80}} & 33.80 & \textbf{\textcolor{red}{21.11}} \\
  \cline{2-9}
  & \multirow{3}{*}{3000} & baseline & 44.44 & 27.49 & 22.72 & 21.84 & 30.07 & 15.42 \\
  & & w/ model pool & - & 29.40 & 23.66 & 23.69 & 29.60 & 15.23 \\
  & & w/ KD & - & 27.98 & 24.67 & 21.78 & \textbf{\textcolor{red}{35.46}} & 19.24 \\
  & & w/ model pool \& KD & - & \textbf{\textcolor{red}{29.98}} & \textbf{\textcolor{red}{25.03}} & \textbf{\textcolor{red}{26.53}} & 33.68 & \textbf{\textcolor{red}{20.96}} \\
  \hline
  \hline
  \end{tabular}
  
  \caption{Ablation study result on CIFAR10 dataset. The baseline is the original gradient matching method \textbf{DC}. We test the performance of the distilled dataset on different models with baseline method, only model pool method, only knowledge distillation method, and model pool \& knowledge distillation method. We set the test model as MLP, LeNet, AlexNet, VGG11 and ResNet18 respectively to test the generalization ability of the distilled dataset. The IPC is set as 1 and 10 respectively and the iteration number is set as 1000, 2000, and 3000 respectively.}
  \label{result}  
\end{table*}

\subsubsection{Knowledge Distillation}

{\Large
\setstretch{1.25}
During the knowledge distillation process, a distilled dataset is used. It's important to choose a teacher model with good performance on this dataset. In our experiment, we selected a $3$-layer ConvNet as our model for knowledge distillation in the baseline method and as our "main model" in the model pool method. After testing, we found that the $3$-layer ConvNet had excellent performance on the distilled dataset (as shown in Table \ref{result}). Therefore, we chose the $3$-layer ConvNet as the teacher model for our knowledge distillation method.
}

\subsection{Result Analysis}\label{section:Result Analysis}

{\Large
\setstretch{1.25}
The experiment results are shown in Table \ref{result}. We test the generalization ability of the distilled dataset on our model pool method and knowledge distillation method together and separately and compare them with the baseline method. 'ConvNet' is the main model in our model pool method, and we test the generalization ability of the distilled dataset on 'MLP', 'LeNet', 'AlexNet', 'VGG11', and 'ResNet18' respectively. We can see that our model pool method and knowledge distillation method together can cause a great improvement in the generalization ability of the distilled dataset compared with the baseline method. 

}

\section{Conclusions}

{\Large
\setstretch{1.25}
This paper studies the problem of performance reduction in cross-architecture tests in dataset distillation. Toward this problem, we propose a novel method called model pool in the dataset distillation process. And we integrate the knowledge distillation method into the training process. Our experiment results show that combining the model pool method and knowledge distillation method can mitigate the performance reduction in cross-architecture tests. We believe that our work can provide a new perspective for future research of dataset distillation.

}

\section{Future Work}

{\Large
\setstretch{1.25}
The experiment results above illustrate that our model pool is more efficient than previous existing methods. In the appendix \ref{sec:appendix}, we have conducted some experiments \ref{model_pool_ablation} to explain the reason why we should choose a main model from the model pool rather than random choosing. However, the probability of choosing a main model is fixed at 0.9 by our prior knowledge. In the future, we can change the different probabilities of the main model automatically to find the best parameter for the generalization ability of the mixed model.

\Large
\setstretch{1.25}
Besides, following the original gradient matching paper \cite{zhao2020dataset}, we set the model depth as 3. However, in deep learning circumstances, this depth may be too shallow compared to the common model nowadays. So in the future, we can make the network deeper, 30 layers or more for instance. We consider the deeper network has a stronger ability, but also a stronger bias, which makes it difficult to generalize the distilled dataset. 

}

\begin{acks}
We thanks Cewu Lu and Yonglu Li for their patient guidance and help.
\end{acks}

\nocite{*}

\bibliographystyle{plain}

\bibliography{ref}

\appendix

\clearpage
\section{Setting of Model Pool}
\label{sec:appendix}

\begin{table*}[p]
  \centering  

  \begin{tabular}{|c|c|c|c|c|c|c|c|}  
  \hline
  \textbf{ipc} & \textbf{iteration} & \textbf{method}  & \textbf{MLP}  & \textbf{LeNet} & \textbf{AlexNet} & \textbf{VGG11} & \textbf{ResNet18}  \\
  \hline
  \multirow{9}{*}{1} & \multirow{3}{*}{1000} & total random & 10.29 & 10.11 & 9.92 & 11.35 & 10.39 \\
  & & w/ main model & 16.27 & 16.48 & 10.34 & 20.08 & 11.42 \\
  & & w/ main model \& similar other models & \textbf{\textcolor{red}{27.24}} & \textbf{\textcolor{red}{20.22}} & \textbf{\textcolor{red}{16.88}} & \textbf{\textcolor{red}{25.11}} & \textbf{\textcolor{red}{17.81}} \\
  \cline{2-8}
  & \multirow{3}{*}{2000} & total random & 10.1 & 10.94 & 10.00 & 13.62 & 10.40 \\
  & & w/ main model & 14.81 & 17.06 & 9.96 & 21.00 & 11.69 \\
  & & w/ main model \& similar other models & \textbf{\textcolor{red}{27.49}} & \textbf{\textcolor{red}{20.88}} & \textbf{\textcolor{red}{20.54}} & \textbf{\textcolor{red}{25.69}} & \textbf{\textcolor{red}{18.65}} \\
  \cline{2-8}
  & \multirow{3}{*}{3000} & total random & 10.00 & 10.80 & 10.04 & 13.60 & 10.41 \\
  & & w/ main model & 10.00 & 10.66 & 9.65 & 19.60 & 11.48 \\
  & & w/ main model \& similar other models & \textbf{\textcolor{red}{27.73}} & \textbf{\textcolor{red}{22.31}} & \textbf{\textcolor{red}{19.47}} & \textbf{\textcolor{red}{25.06}} & \textbf{\textcolor{red}{18.73}} \\
  \hline
  \hline
  \multirow{9}{*}{10} & \multirow{3}{*}{1000} & total random & 24.05 & 16.36 & 16.10 & 23.15 & 10.91 \\
  & & w/ main model & 27.63 & 19.75 & 20.12 & 30.29 & 12.86 \\
  & & w/ main model \& similar other models & \textbf{\textcolor{red}{29.37}} & \textbf{\textcolor{red}{25.82}} & \textbf{\textcolor{red}{23.70}} & \textbf{\textcolor{red}{34.97}} & \textbf{\textcolor{red}{18.34}} \\
  \cline{2-8}
  & \multirow{3}{*}{2000} & total random & 25.88 & 17.93 & 13.74 & 24.62 & 10.96 \\
  & & w/ main model & 27.74 & 21.03 & 20.58 & 31.28 & 13.44 \\
  & & w/ main model \& similar other models & \textbf{\textcolor{red}{30.48}} & \textbf{\textcolor{red}{25.76}} & \textbf{\textcolor{red}{27.24}} & \textbf{\textcolor{red}{34.99}} & \textbf{\textcolor{red}{18.69}} \\
  \cline{2-8}
  & \multirow{3}{*}{3000} & total random & 24.62 & 17.12 & 12.57 & 25.37 & 11.21 \\
  & & w/ main model & 28.59 & 21.15 & 20.92 & 31.57 & 13.62 \\
  & & w/ main model \& similar other models & \textbf{\textcolor{red}{30.78}} & \textbf{\textcolor{red}{25.07}} & \textbf{\textcolor{red}{28.26}} & \textbf{\textcolor{red}{35.50}} & \textbf{\textcolor{red}{19.33}} \\
  \hline
  \hline
  \end{tabular}
  
  \caption{Ablation results on different model pool method settings. Method 'total random' means putting ConvNet, LeNet, AlexNet and VGG11 into the model pool with the same probability of 0.25. Method 'w/ main model' means that putting ConvNet, LeNet, AlexNet and VGG11 into the model pool, and setting ConvNet as the main model with probability 0.9 and setting LeNet, AlexNet and VGG11 with probability $\frac{0.1}{3}$. Method 'w/ main model \& similar other models' means that using the same setting as Table $\ref{MainConvNet}$ and Table $\ref{RandomConvNet}$.}
  \label{model_pool_ablation}  
\end{table*}
{\Large
\setstretch{1.25}
In this part, we will show the ablation result of different model pool method settings. We will show the result of three settings. The first one is that we use the original definition of model pool, which means that we don't use the main model and we choose various models in the model pool. In this setting, we put ConvNet, LeNet, AlexNet and VGG11 into the model pool with the same probability of 0.25. The second one is that we use the main model and we choose various models for the other models in the model pool. In this setting, we still put ConvNet, LeNet, AlexNet and VGG11 into the model pool, and we set ConvNet as the main model with probability 0.9 and set LeNet, AlexNet and VGG11 with probability $\frac{0.1}{3}$. The third one is that we use the main model and we choose similar models for the other models in the model pool. In this setting, we use the same setting as Table $\ref{MainConvNet}$ and Table $\ref{RandomConvNet}$.

The ablation result is shown in Table $\ref{model_pool_ablation}$. As we can see in the table, the third setting can achieve the best performance in all the cases and it's largely better than the first setting and the second setting. Also, we will say some accuracies are equal to 10 and even lower than 10 in the first setting and the second setting. That is because the model pool is too various and the parameters are divergent in the training process. That's why we use the third setting, which has a main model and similar other models, to experiment with the main part of this paper.

}

\end{document}